# Intelligent Hybrid Man-Machine Translation Quality Estimation

**Ibrahim Sabek** [1], **Noha A. Yousri** [1], **Nagwa Elmakky** [1] and **Mona Habib** [2]

(1) Comp. Sys. and Eng. Department, Alexandria University, Egypt
(2) Advanced Technology Labs (ATL) Cairo, Microsoft Research, Egypt

```
{ibrahim.sabek, noha.yousri, nagwa.el-mekyky}@alexu.edu.eg,
                   mona.habib@microsoft.com
```

## Abstract

Inferring evaluation scores based on human judgments is invaluable compared to using current evaluation metrics which are not suitable for real-time applications e.g. post-editing. However, these judgments are much more expensive to collect especially from expert translators, compared to evaluation based on indicators contrasting source and translation texts. This work introduces a novel approach for quality estimation by combining learnt confidence scores from a probabilistic inference model based on human judgments, with selective linguistic features-based scores, where the proposed inference model infers the credibility of given human ranks to solve the **scarcity** and **inconsistency** issues of human judgments. Experimental results, using challenging language-pairs, demonstrate improvement in correlation with human judgments over traditional evaluation metrics.

## 1 Introduction

Quality Estimation (QE) has grasped the attention of professional readers and translators as the main users of Machine Translation (MT), because it provides a quality indicator for unseen translated sentences at various granularity levels. Usually, those users cannot understand the source language, and have difficulties to detect the incorrect translation of ambiguous words, the incorrect assignment of semantic roles in a sentence, or a reference to incorrect antecedent.

Translations with mentioned problems have adequacy problems and are usually produced by SMT systems that tend to provide fluent translations more than adequate ones. A lot of research work in QE e.g. (Aziz, 2011), provides solutions to measure the fluency and adequacy of a sentence, based on human judgments and a number of translation quality indicators from the source and target texts using a variety of simple frequency and linguistic information. However, there are two main issues in dealing with human judgments. First, human judgments are too expensive to collect, so they are usually *scarce*. Second, sometimes different human judgments disagree for the same sentence. Thus, it is required to have a model that learns the uncertainties in relevance between human judgments based on these discrepancies. Moreover, having a model that further predicts human-like scores in the absence of real ones, would be greatly appreciated.

This work introduces a novel approach for quality estimation by combining learnt confidence scores from a probabilistic inference model based on human judgments, with selective linguistic features-based scores, where the proposed inference model infers the credibility of given human ranks to solve the **scarcity** and **inconsistency** issues of human judgments. The inference model addresses the high variations among human judgments by learning uncertainties in human scores and identifying bad judgments to be discarded or re-examined. Experiments with French-English and Spanish-English translations, show that the proposed approach provides more accurate estimation scores for new translations in terms of better correlation with human judgments. We believe this could be a promising direction, and with rigorous testing and tuning, a potential candidate for the real-world applications.

The rest of the paper is organized as follows: Section 2 summarizes the related work in the field of quality estimation for MT. In Section 3, we present the hybrid quality estimation approach, and give the details of the probabilistic inference model. In Section 4, we give an overview of the system architecture. In Section 5, we present our experiments and results.

## 2 Related Work

Sentence-level quality estimation (QE) - also called *confidence estimation* - can be divided into two categories: (1) estimating general quality score by automatic evaluation metrics like BLEU (Papineni, 2002) and METEOR (Lavie, 2005), and (2) estimating scores for post-editing effort based on users ranking and without reference translations like in (Soricut, 2010; He, 2010; Specia, 2011; Aziz, 2011).

Estimating translation quality on the sentence level was introduced earlier using evaluation metrics based on string-based comparisons between candidate translations and reference translations, where human judgment aspects such as translation adequacy and fluency are considered. Both BLEU (Papineni, 2002) and NIST (Doddington, 2002) evaluate a candidate translation by counting the number of n-grams shared with one or more reference translations in the corpus, with NIST additionally using frequency information to favor some n-grams to others. Also, (Melamed, 2003) introduced General Text Matcher (GTM) using measures of precision, recall and F-measure with graphical interpretation to evaluate the translation quality. (Lavie, 2005) proposed METEOR based on an explicit word-to-word matching between candidate translation and one or more reference translations. Real-time applications that rely on scoring translations as a prerequisite (e.g. Post-editing translation or refining search engine content), however, require more than adequacy and fluency from the human point of view. These applications need also effective scoring metrics on the sentence level. In this direction, one could benefit from efficient scoring techniques used in machine translation systems (e.g. Phillips, 2011), to include more specific linguistic features (data-driven features). Hybrid evaluation techniques, that integrate human and data-driven/linguistic features, have been previously sought to obtain a better accuracy. However, they come at a prohibitively high cost, mostly in the form of extensive sentences annotation and labeling for different sentence parts. (Specia, 2011) uses explicit human annotations for each instance as features to estimate the translation quality.

Quality estimation (QE) has been introduced from a new perspective by evaluating new translations based on user rankings obtained for similar previously stored translations instead of using reference translations. It aims at removing the need for reference translations and generally uses machine learning techniques to predict quality scores. This topic has been a value-added service from the user perspective as follows: 1) It can decide the suitability of translation to be published as is (Soricut, 2010). 2) It can filter out sentences that need high effort for post-editing (Specia, 2011). 3) It can select the best translation among options from multiple MT and/or translation memory systems (He, 2010). 4) It can inform readers of the target language about the reliability of translations (Aziz, 2011). Although this topic is still very recent, it shows promising results and opens the way for using human judgments as they are without explicit annotation or labeling for each sentence part. Repositories for human judgments have been thus available for evaluating quality estimation techniques. The shared translation task of WMT (Callison-Burch, 2007) is considered one of the main resources that supply human judgments/votes for given sets of translations in different languages. The presence of such repository is motivating to use in learning the behavior of human voting, compared to other evaluation techniques.

Various linguistic features were proposed to score translation instances. Abstract linguistic features were proposed to evaluate machine translation as a classification problem (Corston-Oliver, 2001). (Amigó, 2006) showed that metrics incorporating deep linguistic information are robust compared with lexical-based metrics. (Shen, 2009) defined feature functions in a practical way to capture linguistic and contextual information in translations. The main observation in these approaches is the greedy nature of integrating all available features which results in low accuracy if there are too many cross-dependent features. (Yang, 2011) provided an engineering solution for selecting the best set of scoring features. (Phillips, 2011) proposed a joint model of SMT and example-based MT based on a selective set of statistical and alignment-based features and showed their superiority over previous systems. The success of alignment models in delivering accurate MT outputs inspired us to exploring them in translation quality estimation.

## 3 Hybrid Quality Estimation Approach

Most quality estimation approaches work by training classifiers using previously assessed translations and a set of weighted features as quality indicators. We differ from that by building an inference model for predicting the credibility of given human ranks as *confidence scores* through a probabilistic model, and then use these scores to weigh the quality indicators.

We can categorize the quality indicators into three categories: 1) *Alignment-based Indicators* (AI), a set of features that measure the word correspondences in the sentence-pair of translations. 2) *Coverage-based Indicators* (CI), a set of features that measure the matching between source and target sentences of a translation. 3) *Frequency-based Indicators* (FI), a set of features that reflect the popularity of source and target sentences of the translation generated by the SMT system. In this paper, we make use of the three set of features which can be extracted as shown by (Phillips, 2011). We focus on the language-independent features that can be extracted generically for any pair of languages, which is a typical scenario for users who use online MT systems to obtain the gist of texts such as professional translators and readers.

The inference model can be trained from observations of a set of ordinal ratings on a user specific scale. Efficient inference is achieved by approximate message passing involving a combination of Expectation Propagation (EP) and Variational Message Passing (VMP). We used the Infer.NET (Minka, 2010) library to perform required computations.

### 3.1 Features

In this work, we use a selective set of features from all categories, namely: Alignment-based, Coverage-based and Frequency-based indicators. These language-independent features are extracted from the datasets provided in the evaluation section. In what follows we describe the set of features proposed in this paper, and list some examples of the already existing features - to check the complete list of features, please refer to (Phillips, 2011).

**AI : Alignment-based Indicators**

- Outside phrase alignment probabilities as shown in (Phillips, 2011).
- Inside phrase alignment probabilities as shown in (Phillips, 2011).
- Uncertainty probability threshold of the phrase alignment score.

**CI : Coverage-based Indicators**

- Ratio of the number of source words covered by the target sentence.
- Ratio of the number of target words covered by the source sentence.

**FI : Frequency-based Indicators**

- Source/Target sentence length.
- Average source/target word length.
- Source/Target sentence occurrences ratio.

For the alignment-based features, the number of n-grams per phrase is a key factor in enhancing the estimation score. Small phrases increase possible aligned source and target units that will be used later to generate features. It will be shown in Section 5 that a significant relative improvement in correlation with human judgments can be obtained by reducing the number of n-grams per phrase. The complete process of extracting features is described in Section 4.

### 3.2 The Probabilistic Inference Model

In designing this model, we start by thinking about the nature of the human judgments/votes. Given a translation, a human vote can be categorized as one of three categories 0 (Bad), 1 (Needs moderate effort) or 2 (Good) that stands for different rankings of the translations. Each voter provides only one rank per translation, and each translation is evaluated by many voters. It is assumed that all votes are equally treated, with no biases, which means that we are confident of the voters experience (an assumption that can later be relaxed, and included in the model). Due to the *scarcity* nature of human assessments, each human rank can be considered to provide a bit of evidence about the quality of the translation. The more human ranks we have, the more we get confident about the translation rank.

The system monitors the change in the number of voters in the system by time. It reflects this change on the confidence of ranks for each translation. Fluctuating between increasing and decreasing the number of voters, would certainly decrease the confidence. Table 1 shows an example on generated confidence scores for different voters changes. It can be seen that the more voters there are, the more confident the ranking model is. For example, 10 voters out of 20 is not enough to be as confident as the case of having 999 voters out of 1000.

| # Prev. voters | # Curr. voters | Confidence score |
|---|---|---|
| 1000 | 999 | 0.9490 |
| 10 | 9 | 0.7929 |
| 20 | 10 | 0.4959 |

Table 1: The effect of changing number of voters on the confidence score

### 3.2.1 Probabilistic Rating Model

Initially, let us assume that inference model receives tuples $(x, y, r)$ of user description $x \in R$, translation description $y \in R$ and ratings $r \in R$. We define user trait variable as $s = u\,x$ where $u$ is a latent user trait variable. Similarly, we define the translation trait variable $t = v\,y$ where $v$ is a latent translation trait variable. Now the probability of rating $r$ is modeled as

$$p(r \mid s, t) = N(r \mid s\,t, \beta^2)$$

where $N$ is a normal distribution, and $\beta$ is the standard deviation of the observation noise. Thus, we adopt a form in which the expected rating of a user to a certain translation is given by the inner product of the user and translation traits. The model parameters to be learned are the variables $u$ and $v$ which determine how users and translations are mapped to the trait space. We represent our prior beliefs about the values of these parameters by independent Gaussian distributions. For example,

$$p(u) = N(\mu_u, \sigma_u^2)$$

and similarly for $p(v)$. We choose this factorizing prior because it reduces memory requirements to two parameters ( a mean and standard deviation ) and it allows us to perform efficient inference as shown later.

### 3.2.2 Adaptation to Ordinal Ranks

A common scenario is that users provide feedback about which translations they like or dislike via an ordinal ranks. These ranks can only be compared, but not subtracted from one another. In addition, each user's interpretation of the scale may be different and the mapping from rank to latent rating may not be linear. We assume that for each user-translation pair for which data is available we observe a rank $l = \{0, 1, 2\}$. We relate the latent rating $r$ to ranks $l$ via a cumulative threshold model (Chu, 2005). For each user $u$, we maintain user-specific thresholds $b_u \in R^{L-1}$ which divide the latent rating axis into $L$ consecutive intervals $(b_{u(i-1)}, b_{u(i)})$ of varying length each of which representing the region in which this user gives the same rank to a translation. Formally, we define a generative model of a ranking as

$$p(l = a \mid b_u, r) =$$
$$\begin{cases} \prod_{i=0}^{a-1} I(r > b_{\widetilde{u(i-1)}}) \prod_{i=a}^{L-1} I(r < b_{\widetilde{u(i-1)}}) & \text{if } 1 < a < L \\ \prod_{i=1}^{L-1} I(r < b_{\widetilde{u(i-1)}}) & \text{if } a = 1 \\ \prod_{i=a}^{L-1} I(r > b_{\widetilde{u(i-1)}}) & \text{if } a = l \end{cases}$$

where

$$p(b_{\widetilde{u(i)}} \mid b_{u(i)}) = N(b_{\widetilde{u(i)}}; b_{u(i)}, \tau^2)$$

and we place an independent Gaussian prior on the thresholds $p(b_{u(i)}) = N(b_{u(i)}; \mu_{u(i)}, \sigma^2)$. The indicator function $I(.)$ is equal to 1 if the proposition in the argument is true and 0 if it is false. Inferring these thresholds for each user allows us to discard extreme or *inconsistent* ranks compared to the expected range of ranks of her.

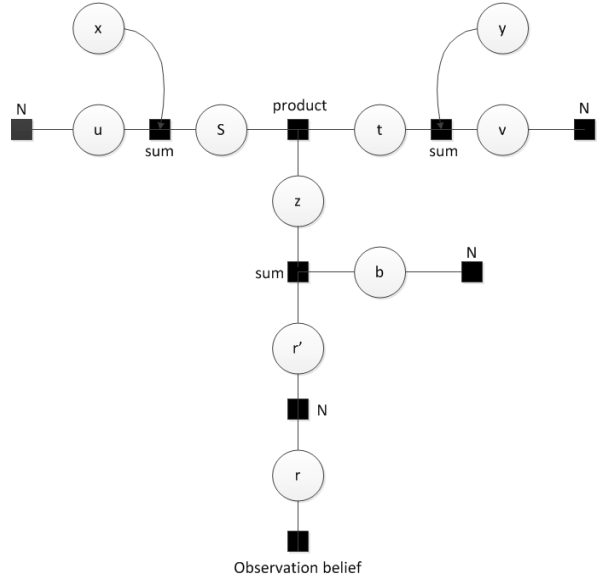

Figure 1: Factor graph of the inference model.

## 3.3 Inference

Given a stream of rating tuples $(x, y, r)$, we train the model in order to learn posterior distributions over the values of the parameters $u$ and $v$. This can be accomplished efficiently by message passing (Minka, 2010). The model described in Section 3.2.1 can be further factorized by introducing some intermediate latent variables $z_k$ to represent the result of the inner product of $s_k$ and $t_k$ where $k$ represents a certain user. Thus, $p(z_k \mid s_k, t_k) = I(z_k = s_k\,t_k)$, Now the latent rating over a set of human ranks is given by

$p(r^\sim | \mathbf{z}, b) = I(r^\sim = \sum_k z_k + b)$. From the probabilistic model in Section 3.2.1, we can estimate that, $p(s_k | u_k, x_k) = I(s_k = u_k x_k)$ and $p(t_k | v_k, y_k) = I(t_k = v_k y_k)$. Therefore the joint distribution of all the variables factorizes as $p(s,t,u,v,z,r^\sim,r | x, y, z) =$

$p(r | r^\sim) \, p(r^\sim | \mathbf{z}, b) \, p(u) \, p(v) \prod_{i=1}^{k} p(z_k | s_k, t_k) \, p(s_k | u_k, x_k) \, p(t_k | v_k, y_k)$.

The factor graph for this model is shown in Figure 1. Factor graph is a bipartite graph with (square) factor nodes corresponding to factors in a function and (circular) variable nodes representing variables in the function. The edges of the graph reveal the dependencies of factors on variables (Loeliger, 1998). Message passing is used to compute the marginal of a joint distribution by assuming a full factorization of the joint distribution. Representing the model in terms of conditional probabilities rather than deriving generative models leads to solving the *scarcity* problem of observed users-translations pairs for training the model.

## 4 System Architecture Overview

In this section, we describe the overall system, mentioning the selected features used for scoring and where human judgments are learnt and incorporated in the final score. Figure 2 shows the block diagram of our system that has four main components:

*Word Alignment Module* uses GIZA++ (Och, 2003) in the offline phase to create an index of source-to-target and target-to-source word alignments from the source-target sentences, stored in the parallel corpus (Callison-Burch, 2007), for further scoring stage.

*Linguistic-based Scoring Module* encodes all possible contiguous phrases from the source and target input translation. Matches are then retrieved from both of source and target corpuses. For each source(target) match, phrase-alignment matrices are built, based on the generated word alignment matrix. The matrices are then used to generate initial evaluation scores via alignment-based and linguistic scoring features. Having source-to-target and target-to-source features set is the key difference from (Phillips, 2011).

*Human Inference Module* runs, during the offline phase, on the humanly judged translations from the parallel corpus that will be used later to extract features. Each translation can be judged by more than one human, and the human can judge many translations. For each translation, maybe there are some extreme judgments that should be discarded. Also, we could have more voters with various judgments later. Thus, this module captures the effect of votes variation and reflects it as learnt confidence scores for this sentence that will be used later to weigh the corresponding features.

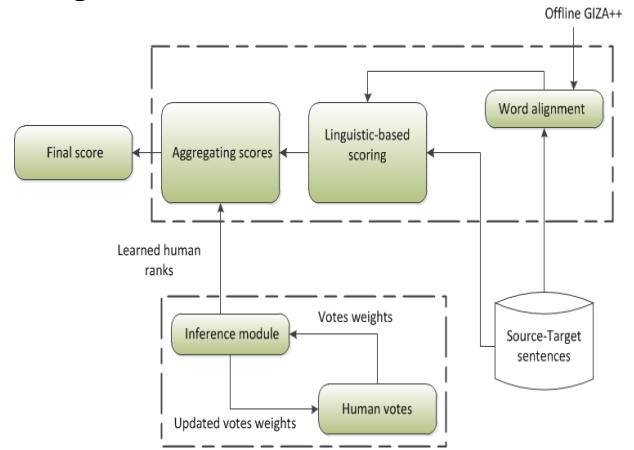

**Figure 2 System Architecture**

*Aggregating Scores Module* adapts the proposed log-linear model in (Phillips, 2011), to aggregate all generated features scores in one final score, by weighing features with the corresponding learnt confidence scores from the Human Inference Module.

## 5 Experiments and Results

In this section, we start by describing the datasets used and human assessment. Then, we compare the system with the state-of-the-art automatic evaluation metrics.

### 5.1 Datasets and Human Judgments

We evaluated our system using the shared task data of the 2007 ACL Workshop on Statistical Machine Translation (Callison-Burch, 2007). Two different challenging language pairs are selected for evaluation (1) French-to-English (2) Spanish-to-English. Extensive human evaluation was carried out per each translation which allows high confidence in the given rank. The shared task data included training, development and testing sets from the Europarl multilingual corpus and the News Commentary data. It is assumed that submissions from different MT systems in the shared task are coming from different translators where each MT system represents a translator. Table 2 shows some statistics about these datasets.

For human annotations, subjective human judgments were collected about the translation quality of each sentence from human annotators on a 1-to-5 scale, we then put these scores on a 0-to-2 scale to apply our approach. Also, automatic evaluation metrics were applied on these translations and the reported correlation values with human judgments can be considered as another way of annotating translations for quality since reference translations are available.

| Dataset | # Snt | # Words | # Distinct words |
|---|---|---|---|
| Fr-to-En Europarl test | 2000 | 53, 981 | 10, 186 |
| Fr-to-En News test | 2007 | 49, 820 | 11, 244 |
| Sp-to-En Europarl test | 2000 | 55, 380 | 10 ,451 |
| Sp-to-En News test | 2007 | 50, 771 | 10, 948 |

**Table 2: French-to-English and Spanish-to-English datasets: number of sentences, number of words and number of distinct words.**

To evaluate the performance of our approach, we compare it with commonly used evaluation metrics according to WMT 07, such as BLEU (Papineni, 2002), GTM (Melamed, 2003), 1-TER, and METEOR (Lavie, 2005). METEOR has special importance because it has been shown to correlate better with the human perception of translation quality in previous research work (Aziz, 2011). The main metric used, to calculate the correlation between human evaluation and the evaluation given by our approach is Spearman's rank correlation coefficient. For fair comparison with reported results in (Callison-Burch, 2007), we use the same simplified Spearman form as follows:

$$\rho = 1 - \frac{6 \sum d_i^2}{n(n^2 - 1)}$$

where $d_i$ is the difference between the rank given by our approach and the leveraged human rank for a certain translation, and $n$ is the number of translations. The possible values of $\rho$ range between 1 (where all translations are ranked in the same manual order) and -1 (where the translations are ranked in the reverse order). The higher correlation we have, the closer to human evaluation we are. For each translation, available human votes were randomly split into 66% for training the inference model, and the remaining votes are used for calculating the correlation during testing. This process was repeated for 5 times to generate different splits, and we calculated the average score.

### 5.2 Performance Evaluation

The accuracy of the proposed approach is directly affected by the number of n-grams per phrase, where choosing smaller phrases increases the evaluation accuracy. 9-gram and 5-gram are suitable configurations for many evaluation metrics with respect to WMT 07 training and test sets. Table 3 shows the significant relative improvement in correlation with human judgments for using 5-grams instead of 9-grams. The improvement in the two test sets are 34.28% for French-to-English and 29.41% for Spanish-to-English.

| Dataset | 5-gram Corr. | 9-gram Corr. |
|---|---|---|
| Fr-to-En News test | **0.942** | 0.7 |
| Sp-to-En News test | **0.885** | 0.68 |

**Table 3 : Correlation with Human for French-to-English and Spanish-to-English datasets: 5-gram and 9-gram.**

However, there is a tradeoff between accuracy and running time as shown in Figure 3. By decreasing the phrase length, the system will incur a latency. If the phrase length is chosen to be less than 5-gram, the system latency will increase by around 200%. So, a 5-gram phrase is a reasonable choice for training and testing sets used for evaluation.

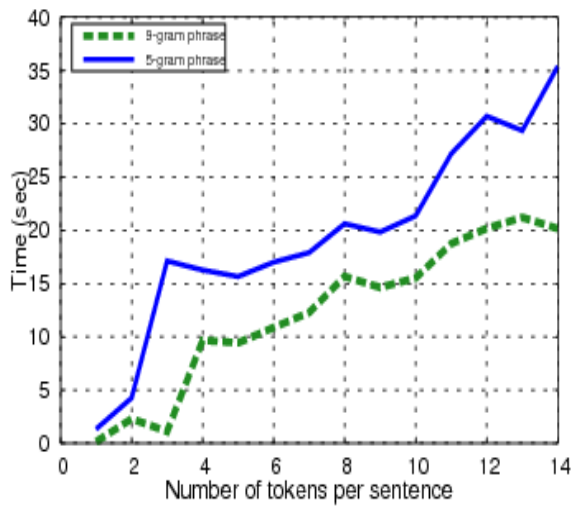

**Figure 3 : The effect of sentence length on the scoring running time.**

Table 4 shows the effect of selecting different combinations of features sets. We observe that the best accuracy (in terms of correlation to human assessment) is achieved when all features are used. This observation is validated by the two language pairs: French-to-English and Spanish-to-English. We can show that probability of ob-taining average correlation value lower than 0.65 is almost zero which ensures a robust lower threshold on the accuracy. In general, including all sets of features results in the best performance.

| Feature sets | Fr-to-En News test | Sp-to-En News test |
|---|---|---|
| AI | 0.88 | 0.65 |
| AI, CI | 0.87 | 0.64 |
| AI, FI | 0.85 | 0.65 |
| AI, CI, FI | **0.94** | **0.88** |

**Table 4: The effect of different combinations of feature sets on correlation with human assessment**

Table 5 shows a correlation comparison of the proposed approach with state-of-the-art evaluation metrics. It could be seen that our scheme shows superiority over other metrics with 19.6% improvement in correlation with human judgment for French-to-English dataset, and 36.1% for Spanish-to-English dataset.

| Metric | Fr-to-En News test | Sp-to-En News test |
|---|---|---|
| METEOR | 0.75 | 0.65 |
| BLEU | 0.78 | 0.35 |
| 1-TER | 0.71 | 0.48 |
| GTM | 0.71 | 0.52 |
| Proposed Sys. | **0.94** | **0.88** |

**Table 5: correlation comparison between our proposed approach and state-of-the-art evaluation metrics**

## 6   Conclusions

This work presents a novel man-machine quality estimation system. The core approach draws from combining a selective set of linguistic features with inferred confidence scores based on given human ranks. An inference model is proposed to predict the human-based scores per each translation while solving the well-known scarcity and inconsistency problems of human judgments. Performance results are promising and motivate us to pursue the system to the point of real deployment. In future, we will expand this work in two directions. Selective features scoring can be used for learning initial prior distributions for each rank, without the need for human judgments. In addition, extensive evaluation will be done using different larger corpora to confirm current results.